\documentclass[10pt, conference, compsocconf]{IEEEtran}
\usepackage{graphicx}
\usepackage{multirow}
\graphicspath{ {./images/} }

\ifCLASSINFOpdf
\else
\fi
\begin{document}
	\IEEEoverridecommandlockouts
	\IEEEpubid{\makebox[\columnwidth]{\textbf{\emph{ACCEPTED BY IEEE BigMM 2020}} \hfill} \hspace{\columnsep}\makebox[\columnwidth]{ }}
%
\title{SGG: Spinbot, Grammarly and GloVe based Fake News Detection}

\author{\IEEEauthorblockN{Akansha Gautam}
\IEEEauthorblockA{Department of Mathematics\\
Indraprastha Institute of Information Technology Delhi\\
New Delhi, India\\
akansha16221@iiitd.ac.in}
\and
\IEEEauthorblockN{Koteswar Rao Jerripothula}
\IEEEauthorblockA{Department of Computer Science and Engineering\\
Indraprastha Institute of Information Technology Delhi\\
New Delhi, India\\
koteswar@iiitd.ac.in}
}


%


\maketitle

\begin{abstract}
Recently, news consumption using online news portals has increased exponentially due to several reasons, such as low cost and easy accessibility. However, such online platforms inadvertently also become the cause of spreading false information across the web. They are being misused quite frequently as a medium to disseminate misinformation and hoaxes. Such malpractices call for a robust automatic fake news detection system that can keep us at bay from such misinformation and hoaxes. We propose a robust yet simple fake news detection system, leveraging the tools for paraphrasing, grammar-checking, and word-embedding. In this paper, we try to the potential of these tools in jointly unearthing the authenticity of a news article. Notably, we leverage Spinbot (for paraphrasing), Grammarly (for grammar-checking), and GloVe (for word-embedding) tools for this purpose. Using these tools, we were able to extract novel features that could yield state-of-the-art results on the Fake News AMT dataset and comparable results on Celebrity datasets when combined with some of the essential features. More importantly, the proposed method is found to be more robust empirically than the existing ones, as revealed in our cross-domain analysis and multi-domain analysis. 

\end{abstract}


\IEEEpeerreviewmaketitle

News consumption using online news portals has increased exponentially in modern society due to cheaper cost, easy accessibility, and rapid dissemination of information. Nevertheless, it also encourages the spread of low-quality information and hoaxes. For example, since the beginning of the COVID-19 pandemic, the distribution of false information has negatively affected individuals and society. Misinformation about the coronavirus ranged from false assertions to harmful health advice. Table~\ref{tab:example_fake_legitimate_covid} shows an example each of COVID-19 related fake and legitimate article. Many times news is new and such newness of news articles, be it legitimate or fake, calls for robustness in the fake news detection problem. That is, fake news detection models should not be domain-dependent.  

\begin{figure}
\centering
\includegraphics[width=80mm,scale=0.5]{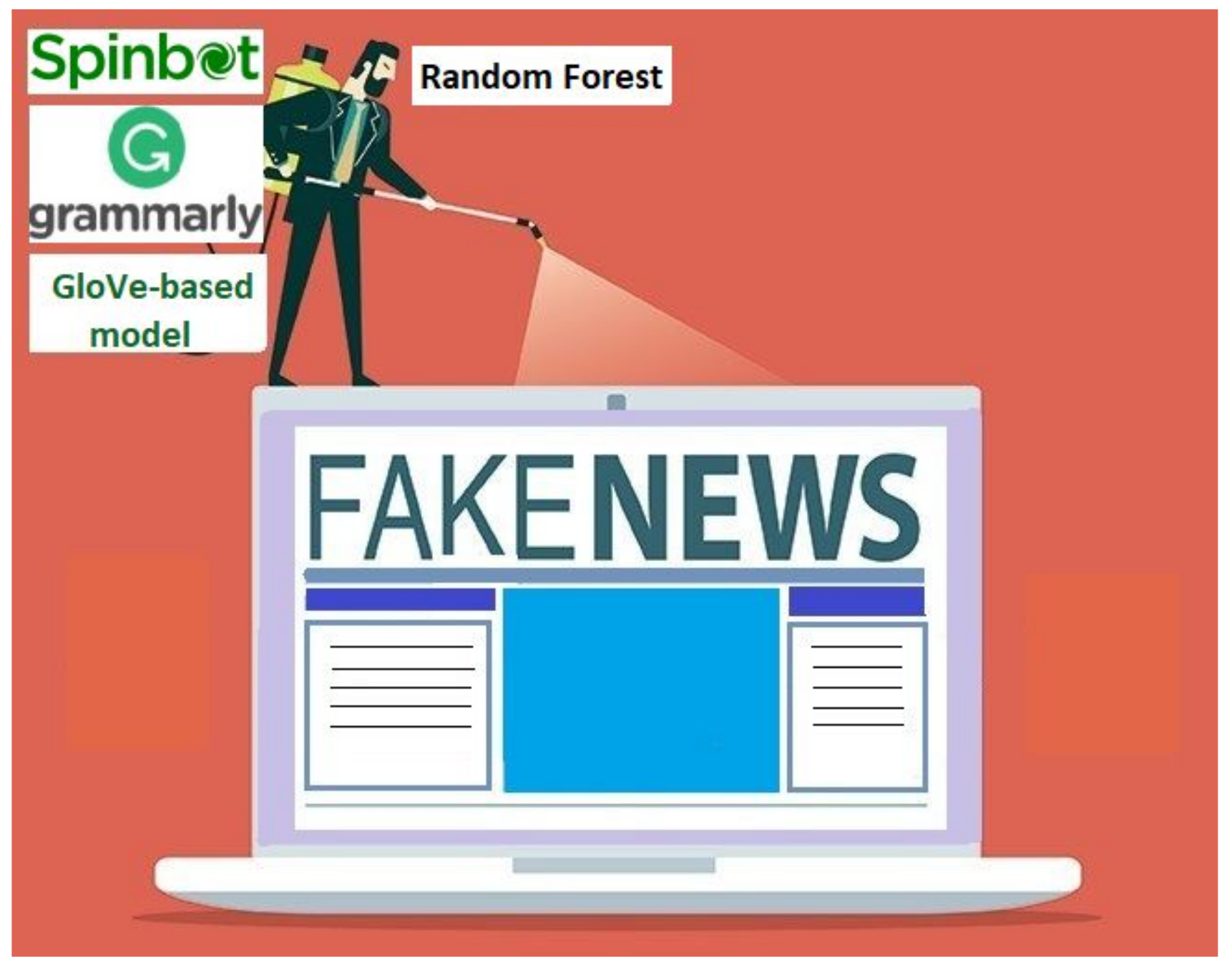}
\caption{Construct novel features using Spinbot, Grammarly, and GloVe-based model to build Random Forest Classifier for the task of fake news detection.}
\label{fig:teaser}
\end{figure}

Considering several such exciting challenges, Fake-news-detection  \cite{Mihalcea:09} has kept attracting several researchers in the last decade. A relatively early study \cite{Ott:11} on {\em deceptive opinions spam} focused on a combination approach that involved computational linguistic and psychological motivating features and $n$-gram features performed better in the detection of ambiguous opinion spam. Various methods have been proposed for automatic fake news detection, covering quite complex applications. Previous work \cite{Rosas:18} introduced two novel fake news datasets, one obtained through a crowdsourcing technique that includes six news domains and another one collected from the web. They also proposed classification models that rely on lexical, syntactic, and semantic information to detect misinformation. To detect satirical news, \cite{Rubin:16} proposed an SVM-based algorithm based on Absurdity, Humor, Grammar, Negative effect, and Punctuation that aids in minimizing the potential deceptive impact of satire. A recent research \cite{NYUTSE:20} report of Science Daily Blog showed that the spread of fake news on social networking sites is a pernicious trend to modern society. It has caused dire implications for the 2020 presidential election. It also manifests that public engagement with false news is higher than with legitimate news from mainstream sources that makes social media a powerful channel for propaganda.   

\begin{table*}
\caption{Example of COVID-19 fake and legitimate news article}
\begin{center}
\begin{tabular}{|p{1.5 cm}|p{15 cm}|}
\hline 
\textbf{News Type} & \textbf{Content} \\ 
\hline
True News & Who can donate plasma for COVID-19?,"In order to donate plasma, a person must meet several criteria. They have to have tested positive for COVID-19, recovered, have no symptoms for 14 days, currently test negative for COVID-19, and have high enough antibody levels in their plasma. A donor and patient must also have compatible blood types. Once plasma is donated, it is screened for other infectious diseases, such as HIV.Each donor produces enough plasma to treat one to three patients. Donating plasma should not weaken the donor's immune system nor make the donor more susceptible to getting reinfected with the virus." \\ 
\hline
Fake News & Due to the recent outbreak for the Coronavirus (COVID-19) the World Health Organization is giving away vaccine kits. Just pay \$4.95 for shipping,"You just need to add water, and the drugs and vaccines are ready to be administered. There are two parts to the kit: one holds pellets containing the chemical machinery that synthesises the end product, and the other holds pellets containing instructions that tell the drug which compound to create. Mix two parts together in a chosen combination, add water, and the treatment is ready."
\\ \hline
\end{tabular}
\label{tab:example_fake_legitimate_covid}
\end{center}
\end{table*}

In this paper, we develop features using Grammarly, Spinbot, and the GloVe-based model. A previous study \cite{Granik:17} has shown that fake news articles often carries a lot of grammatical mistakes. We use Grammarly to extract such features to identify counterfeit items. Spinbot is a paraphrasing tool that is used to make the article content look scholarly. We use an innovative technique to extract sentiment features based on GloVe and K-Means model. GloVe model has shown successful results in representing the meaning of textual data into a feature vector. We use GloVe to transform articles into feature vectors and then apply K-Means to our feature vectors set, where each cluster encapsulates similar words together. After feature extraction, we rely on machine learning algorithms to separate the two categories of news articles. Our proposed methods help correctly identify the fake and legitimate COVID-19 news articles shown in table~\ref{tab:example_fake_legitimate_covid}. Also, it outperforms the previously established works in identifying the misinformation across the web.     

Figure~\ref{fig:teaser} presents our novel approach to a shared task aimed at detecting whether a new article is fake or not. We develop innovative features based on Spinbot, Grammarly, and Glove-based models. We also utilize other essential features to build the classifier for the task of counterfeit news detection. We conduct several experiments to robustly identify linguistic properties that are predominately present in the false news content. 

\section{Related Works}
Detection of misinformation has become an emerging research topic that is attracting the general public and researchers. In recent years, there is a requirement of a practical approach for the success of the fake news detection problem, which has caused a significant challenge to modern society. Each research paper embraces its ideas related to the different strategies to solve this problem. A substantial group of studies fabricates the concept of developing machine learning classifiers \cite{goyalgait} to automatically detect the misinformation using a variety of news characteristics. Some prior studies show that the performance of fake news deception model improves when using features based on the linguistic characteristics derived from the text content of news stories. Previous work \cite{Castillo:11} has shown that newsworthy articles tend to contain URLs and to have deep propagation trees. They built a classification model based on a set of linguistic features such as special characters, sentiment positive/negative words, emojis, etc. to determine information credibility on twitter. Context-Free Grammar(CFG) parse trees \cite{Feng:12} based on shallow lexical-syntactic features were applied to build the detection model. Lexicon patterns and part-of-speed tags \cite{Qazvinian:11} were also used. Named entities, retweet ratio, and clue keywords \cite{Takahashi:12} were analyzed for the identification of rumor tweets on twitter. Another study used linguistic content such as swear words, emotion words, and pronouns \cite{Gupta:14} to assess the credibility of a tweet. Language style and source reliability of articles \cite{Popat:16} were also used in reporting the claim to determine its likelihood. To differentiate fake news from legitimate news, lexical, syntactic, and semantical \cite{Rosas:18} information was used. 

A group of other approaches builds fake news detection model based on temporal-linguistic features. News entities \cite{Shu:18} are categorized into content, social, and temporal dimensions reveals mutual relations and dependencies that were used to perform fake news dissemination. A classifier \cite{Zhao:15} was build based on inquiry phrases extracted from user comments. Prior work \cite{Buntain:2017} has shown that leveraging non-expert, crowdsourced workers provides a useful way to detect fake news in popular Twitter threads. The comprehensive technique of identifying fake news on social media \cite{Shu:19} included fake news characterizations on psychology and social theories. Bimodal variational autoencoder with a binary classifier \cite{Khattar:19} based on multimodal (textual+visual)\cite{10.1007/978-3-319-46478-7_12,shah2017multimodal,8290832} information helped classify posts as fake or not. The work of \cite{Rashkin:17} depicted The significance of stylistic cues was described in determining the truthfulness of text. All scraps that contain propaganda techniques and their types \cite{Martino:19} were detected to design a method that performs fine-grained analysis of texts. Another group of researchers \cite{shivangi:19} exploited both the textual and visual features of an article. In \cite{parikh:19}, the framework was proposed based on information captured from an image\cite{8269367,8099896}, processed, and compared the information with the trusted party was used to detect tampered or photoshoot tweets.     
\begin{figure*}
\centering
\includegraphics[width=1\linewidth]{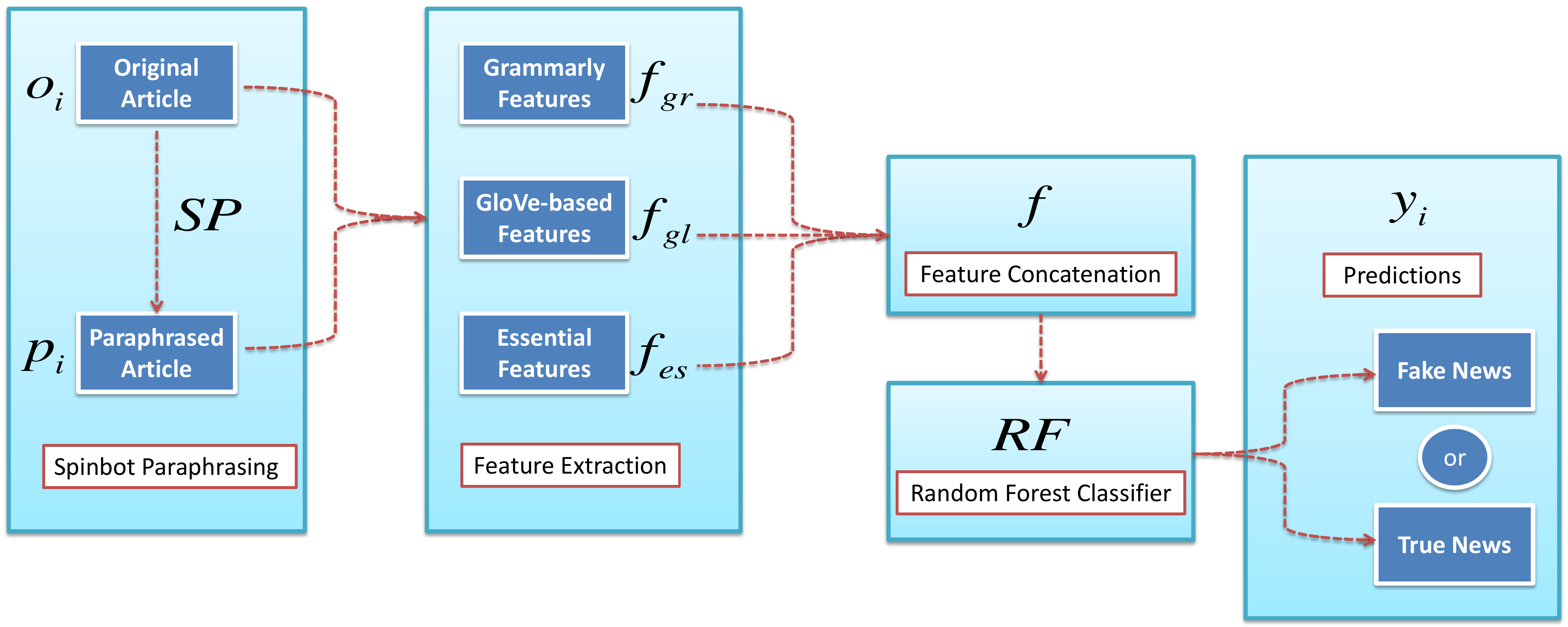}
\caption{Flowchart shows how the features are extracted from the news article and then fed to the classifier}\label{fig:overview}
\end{figure*}
In this paper, we present the extraction of novel features based on Grammarly, Spinbot, and the GloVe-based model. To the best of our knowledge, no previous work has used this kind of approach in extracting features for fake news detection \cite{singhal2019spotfake,singhal2020spotfake+}.            

\section{Proposed Method}
This section precisely describes how we utilize the Spinbot, Grammarly,  and GloVe tools to develop a robust yet simple fake news detection system. We give overview of our proposed approach in the Figure~\ref{fig:overview}. First of all, we paraphrase the original article (denoted as $o_i$) and obtain a paraphrased article (denoted as $p_i$). Then, we extract Grammarly features for both the original article and the paraphrased article. Let's denote such a feature extraction function as $f_{gr}(\cdot)$. Similarly, we obtain GloVe based features (denoted as $f_{gl}(\cdot)$) for both original and paraphrased articles. In addition to these proposed features, we also extract some essential features ($f_{es}(\cdot)$) like TF-IDF, emotion, lexical, and so on, again, for both original and paraphrased articles. We concatenate \cite{7484309} these features into $f$ as shown below:
\begin{equation}
f(o_i,p_i)=f_{gr}(o_i)\cdot f_{gl}(o_i)\cdot f_{es}(o_i)\cdot f_{gr}(p_i)\cdot f_{gl}(p_i)\cdot f_{es}(p_i)
\end{equation}
which is then fed to the random forest classifier (denoted by $RF\big(\cdot\big)$ function) to output the prediction $y_i$, fake news or real news, as shown below:
\begin{equation}
    y_i=RF\big(f(o_i,p_i)\big).
\end{equation}
where we essentially predict the authenticity of a given original article and its paraphrased article using a $RF$ model learned through training on labeled data. Let us now discuss each of these steps in greater detail one-by-one.

\subsection{Spinbot Paraphrasing}
We use paraphrasing to obtain a paraphrased content that looks scholarly. Assuming persons who spread fake news are not so scholarly, such paraphrasing serves two purposes in the false news detection. If they did not use such tools, the paraphrased article would be too distant from the original one, in terms of the features we develop. Also, if they did use it, they will copy-paste the paraphrased article, leading to too much closeness. Either way, fake articles will be distinguishable from real articles that may lie somewhere in-between, neither too close nor too distant. They would not be too close because such automated tools may not completely capture the writer's real intention; so, the writer will change it a bit. 

We use Spinbot\footnote{https://spinbot.com/} tool for this paraphrasing task. Spinbot is a free automatic article spinning tool that rewrites the human-readable text into additional, readable text and helps in solving problems in optimizing content generation. Spinbot allows us to copy text and paste it into a text box. After selecting a couple of basic options and confirming that one is human, users receive an output of a paraphrased version of the original text entered in the text box below. This tool is vibrant and sophisticated; it makes the content look very scholarly. It rephrases the text by spinning textual data of length up to 10,000 characters in a single go. Spinbot replaces many words with their synonyms to come up with plausibly ``new" content. Spinbot provides unique, quality textual content that can quickly acquire legitimate web visibility in terms of human readership and search engine exposure. Let's denote this tool as $SP$ function that takes $o_i$ as input and outputs $p_i$, as mentioned in Fig.~\ref{fig:overview} and below.  

\begin{equation}
    p_i=SP(o_i)
\end{equation}

\begin{table*}
\centering
\caption{Grammarly features with their descriptions}
\begin{tabular}{ |p{4 cm} | p{12 cm}|  }
\hline \textbf{Grammarly Feature} & \textbf{Description} \\ 
\hline Contextual Spelling & This Grammarly feature helps identify misspelled words in the text document and suggests proper spellings related to misspelled words. \\ 
\hline Grammar & This feature simplifies the writing process by detecting grammar mistakes that include inappropriate use of articles, verb tenses, and misplaced modifiers. \\ 
\hline Sentence Structure & There are four sentence structures present in the English literature: simple sentences, complex sentences, compound sentences, and complex-compound sentences. This feature checks the proper usage of sentence structure in articles as it detects and prompts the suggestions for correcting missing verbs, faulty parallelism, and incorrect adverb placement. \\ 
\hline Punctuation & Misplaced punctuation changes the meaning of the sentences at a higher level. This feature spots misplaced punctuation and makes real-time suggestions to correct them. \\ 
\hline Style & There are four major writing styles: expository, descriptive, persuasive, and narrative. The style feature analyzes the article to improve the communicating method with readers. \\ 
\hline Vocabulary Enhancement & Grammarly flags vague and redundant words that reduce the value of the content and suggests more engaging synonymous to add variety to articles. \\ 
\hline Plagiarism & Grammarly's plagiarism checking tool checks 16 billion other articles across the web to correctly identify and cite text that is not 100\% original. \\ 
\hline Word Count & This feature provides general metrics that include the characters count, word count, and sentence count of the writings. It also renders the time required in reading and speaking the article. \\ 
\hline Readability & These features indicate text understandability. It computes words and sentences length. It also enumerates the Readability score that measures the reader's likeliness to be understood.  \\ 
\hline Vocabulary & It measures vocabulary depth by identifying words that are not present in the 5,000 most common English words. It also measures the unique words by calculating the number of individual words. \\ 
\hline
\end{tabular}
\label{fig:grammarly-features-description}
\end{table*}

\subsection {Grammarly Features Extraction} 
The first set of features we extract are mistakes-based, assuming the English writing of the persons who spread fake news is not that good. For capturing these mistakes, we use Grammarly. Grammarly is a proofreading application that uses artificial intelligence and natural language processing in developing tools to detect mistakes in grammar, spelling, punctuation, word choice, and style. It is a comprehensive writing tool that provides real-time suggestions to correct these errors. Grammarly checks writings on various aspects such as tone detection, confident language, politeness, formality level, and inclusive language before making a decision. Grammarly scans the document with an AI assistant, automatically detects mistakes, suggests corrections, and shows the rationale behind those suggestions. Interestingly, it can detect up to 250 types of errors. Table~\ref{fig:grammarly-features-description}  shows Grammarly's significant features with their description \cite{grammarly:19}.   Table~\ref{fig:grammarly-features-comparison} shows the Grammarly Features extracted from fake and legitimate news articles shown in Table~\ref{tab:example_fake_legitimate}. There are numerous features for which the values are very different for these examples, thanks to the assumption made.

\begin{table}
\centering
\caption{This table shows Grammarly features extracted for fake and legitimate news articles mentioned in Table~\ref{tab:example_fake_legitimate} where $A$ signifies fake news and $B$ signifies legitimate news}
\resizebox{0.45\textwidth}{!}{%
\begin{tabular}{ | p{3 cm}|p{2 cm} | p{2 cm} | }
\hline \textbf{Grammarly Features} & \textbf{News A}  & \textbf{News B}  \\ \hline
{\bf Overall Score} & 74 & 97 \\
{\bf Correctness Alerts} & 7 & 0 \\
{\bf Clarity} & mostly clear & a bit unclear \\
{\bf Clarity Alerts} & 2 & 2  \\
{\bf Engagement} & engaging & very engaging \\
{\bf Engagement Alerts} & 1 & 0  \\
{\bf Delivery} & just right & just right \\
{\bf All alerts} & 10 & 2  \\
{\bf Characters} & 562 & 476  \\
{\bf Words} & 106 & 71  \\
{\bf Sentences} & 5 & 4  \\
{\bf Reading Time} & 25 sec & 17 sec  \\
{\bf Speaking Time} & 48 sec & 32 sec  \\
{\bf Word Length} & 4.3 & 5.6  \\
{\bf Sentence Length} & 21.2 & 17.8  \\
{\bf Readability Score} & 72 & 58  \\
{\bf Unique Words} & 63 & 79  \\
{\bf Rare Words} & 18 & 29  \\
{\bf Plagiarism} & 0 & 90  \\
\hline
\end{tabular}}
\label{fig:grammarly-features-comparison}
\end{table}

\begin{figure*}
\centering
\includegraphics[width=\textwidth]{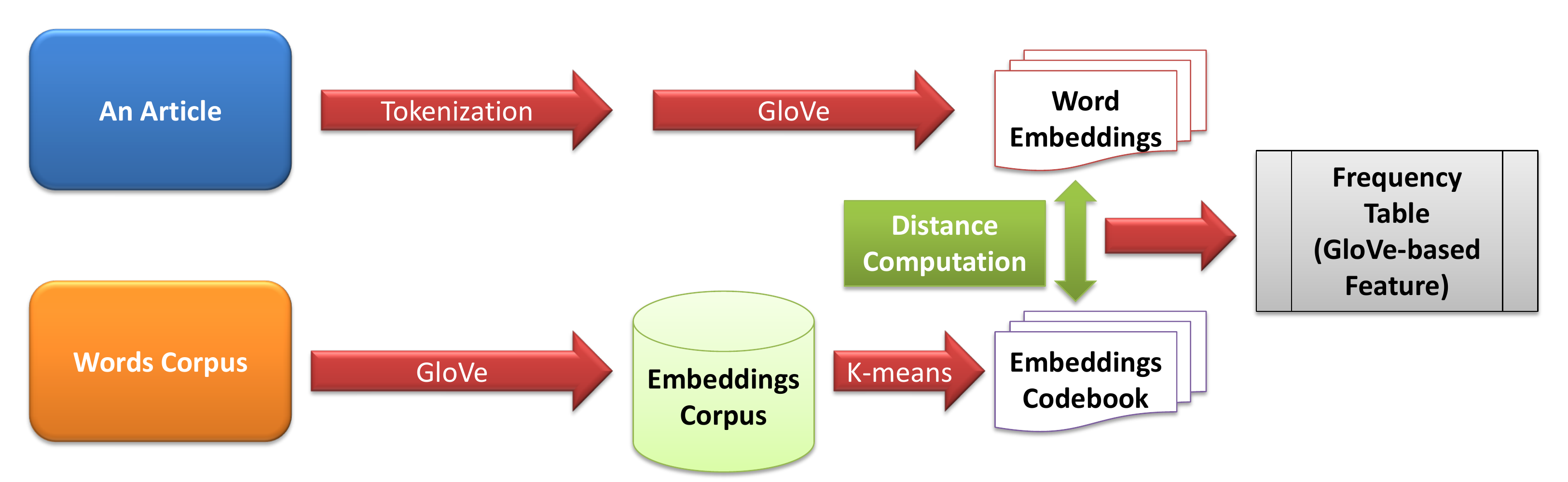}
\caption{Glove-based Feature Generation}\label{fig:accumulator_vector_generate}
\end{figure*}

\subsection{Glove-based Feature Extraction}
	
GloVe embedding is a popular word embedding algorithm. Given the word, it can output a numeric feature vector of $n$ dimensions. Since we need to classify articles, not words, we use a codebook approach to represent an article, as explained below.

We first build a corpus using different words present in a domain. Let us say there is a total of $m$ words in the corpus. We use Glove to convert this corpus into an embeddings corpus, which can be seen as a matrix of dimensions $m\times n$, stacking all the feature vectors vertically. We then apply k-means \cite{macqueen1967some,7025663,7457899,jerripothula2017co,jerripothula2015group} on embeddings corpus to divide the matrix into $k$ clusters of feature vectors. We compute the means of feature vectors in different clusters and assign them as codes in our embeddings codebook. 
The idea is to develop a frequency-table depending upon how many times different words in an article fall close to different codes in our codebook. This frequency-table of the article is what we call as GloVe-based feature. Fig.~\ref{fig:accumulator_vector_generate} depicts our entire approach of generating the GloVe-based feature.

\subsection{Essential Features Extraction}
Our essential features can be classified into three categories: basic, TF-IDF, and emotion. Let us discuss each of them one-by-one. 

\subsubsection{Basic} We extract basic features set consisting of 7 types of lexical features includes Unique word count, Stopword count, URL count, Mean word length, Hashtag count, Numeric count, and Uppercase count for each news article.  

\subsubsection{TF-IDF} We extract unigrams and bigrams derived from the TF-IDF representation of each news article. Term Frequency Inverse Document Frequency (TF-IDF) \cite{Jones:04} gives us information on term frequency through the proportion of inverse document frequency. Words with a small-term rate in each document but have high possibility to appear in records with similar topics will have higher TF-IDF, while words like function words though frequently appear in every report, will have low TF-IDF because of lower inverse document frequency. 

\subsubsection{Emotion Lexicon} We perform sentiment analysis on the corpus of the news articles using Textblob \cite{textblob:18}. Sentiment analysis \cite{9153789} determines the attitude or the emotion of the writer, i.e., whether it is positive, neutral, or negative. The {\em sentiment} function of Textblob returns two properties, polarity, and subjectivity. Polarity returns float value, which lies in the range of [-1,1] where the ``1" stands for a positive statement, and ``-1" stands for a negative comment. Subjective sentences refer to personal opinion, emotion, or judgment, whereas objective refers to factual information. Subjectivity also returns a float number, which lies in the range of [0,1]. 

\subsection{Feature Concatenation and Predictive Modeling}
Different features obtained for both the original and the paraphrased articles are all concatenated into a feature vector. We apply $5$-fold cross-validation to develop our random forest classifier for predicting whether an article is fake or real. Random forest is an ensemble tree-based learning algorithm that fits decision trees on the sub-samples of data and then aggregates the accuracy and controls over-fitting.

\section{Experimental Results}
In this section, we first describe different datasets and how we evaluate our approach. Then, we discuss different results we obtain by evaluating our method on these datasets while comparing them with the prior arts on these datasets.

\subsection{Datasets and Evaluation}

Our experiments are conducted on two publicly available benchmark datasets of fake news detection \cite{Rosas:18}, namely {\em Celebrity} and {\em FakeNewsAMT}. The FakeNewsAMT is collected by combining manual and crowdsourcing annotation efforts, including a corpus of 480 news articles, incorporating six news domains (i.e., Technology, Business, Education, Sports, Politics, and Entertainment). The news of the FakeNewsAMT dataset was obtained from various mainstream websites in the US, such as ABCNews, CNN, USAToday, NewYorkTimes, FoxNews, Bloomberg, and CNET, among others. Table~\ref{tab:example_fake_legitimate} shows examples of fake and legitimate news articles drawn from FakeNewsAMT dataset \cite{Rosas:18}. The Celebrity dataset contains 500 celebrity news articles, collected directly from the web. The news of the Celebrity dataset was obtained predominately from online magazines such as Entertainment Weekly, People Magazine, Radar Online, and other entertainment-oriented publications. The distribution of both the categories of news (fake and legitimate) are evenly distributed in both the datasets. We have extended these two datasets with our rich features.  

\begin{table*}
\caption{Example of fake and legitimate news article}
\begin{center}
\begin{tabular}{|p{1.5 cm}|p{2 cm}|p{12 cm}|}
\hline 
\textbf{Article} & \textbf{Type} & \textbf{Content} \\ 
\hline
$A$ & Fake News & Super Mario Run to leave app store
The once popular Super Mario Run will be taken out of the Google play and apple app store on Friday.  Nintendo says that shortly after its release the public stopped downloading the game when current players had spread the word that in order to play the entire game you had to make an in app purchase. Nintendo and Mario fans are appalled that Nintendo would release a game for free and then charge to play it. Nintendo says they will take the game back to the drawing board, and try and release a free version at a later time.
\\ \hline
$B$ & Legitimate News & How does nutrition affect children's school performance? As politicians debate spending and cuts in President Donald Trump's proposed budget, there have been questions about the effects of nutrition programs for kids. From before birth and through the school years, there are decades-old food programs designed to make sure children won't go hungry. Experts agree that the nutrition provided to millions of children through school meal programs is invaluable for their health.
\\ \hline
\end{tabular}
\label{tab:example_fake_legitimate}
\end{center}
\end{table*}

We analyze our method's performance using measurement metrics like Confusion Matrix and Accuracy. 

\begin{figure}
  \centering
  \begin{minipage}{0.3\textwidth}
    \includegraphics[width=\textwidth]{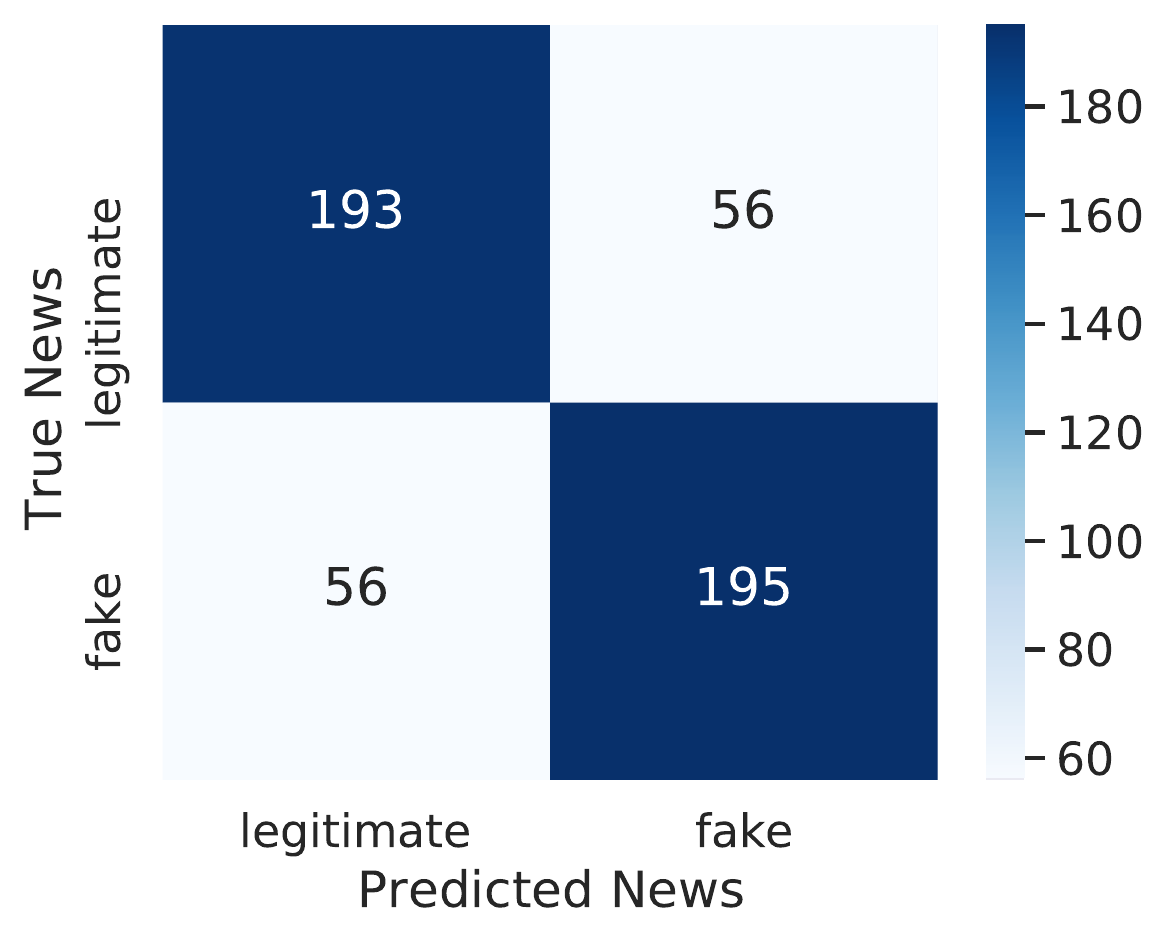}
    \caption{Confusion matrix for the Celebrity dataset} \label{fig:confusion_matrix_celeb}
  \end{minipage}
  \hfill
  \begin{minipage}{0.3\textwidth}
    \includegraphics[width=\textwidth]{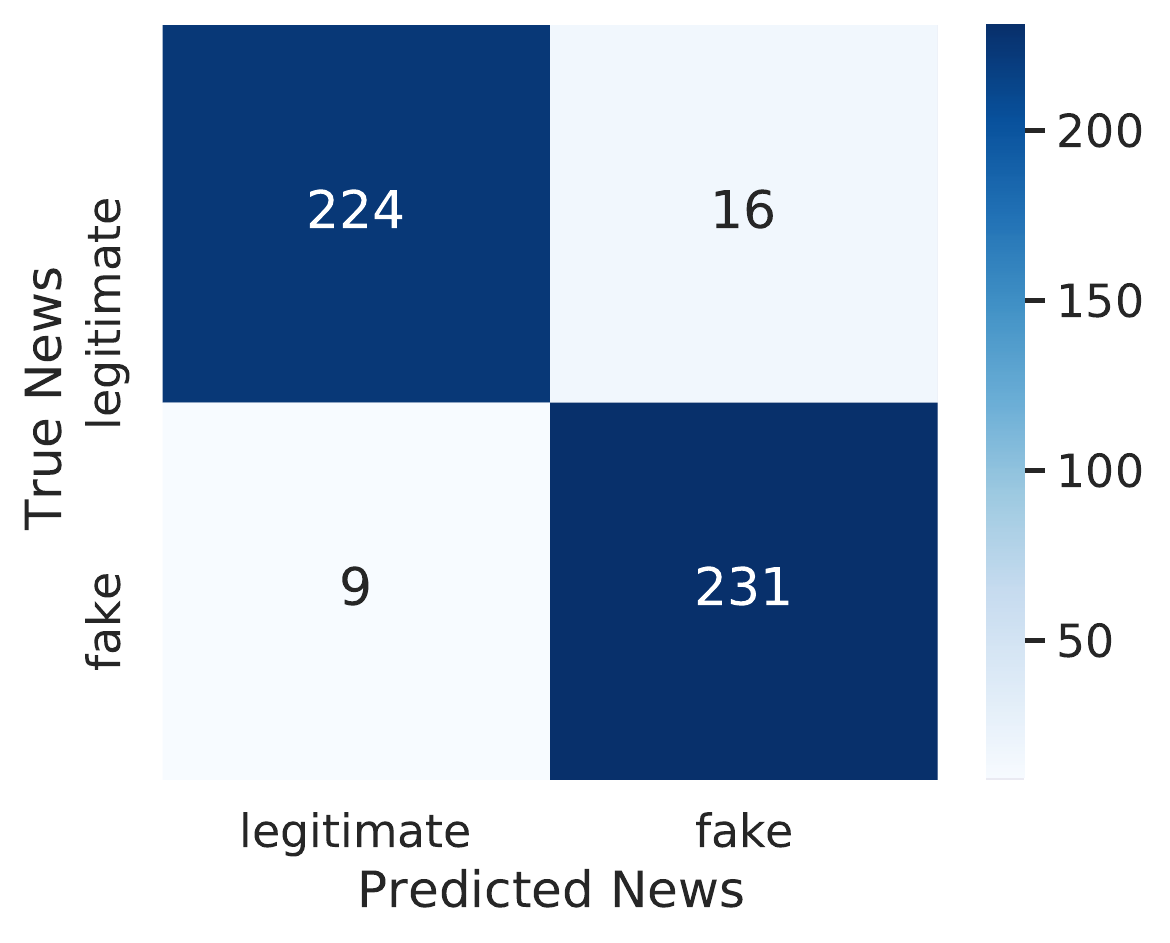}
      \caption{Confusion matrix for the FakeNewsAMT dataset} \label{fig:confusionMatrix_amt}
  \end{minipage}
\end{figure}

\begin{table}
\centering
\caption{Classification results for Celebrity dataset}
\label{tab:results_celebrity}
\resizebox{0.25\textwidth}{!}{%
\begin{tabular}{ |l|l| }
\hline \bf Features (\# features) & \bf Accuracy \\ \hline
Glove based feature vector (200)  & 0.71  \\
Basic (14)                & 0.60 \\
Grammarly (19)            & 0.63 \\
TF-IDF (100)              & 0.76 \\
Emotion(4)                & 0.57 \\
\hline All features (337) & 0.78 \\
\hline
\end{tabular}}
\end{table}

\begin{table}
\centering
\caption{Classification results for FakeNewsAMT dataset}
\resizebox{0.25\textwidth}{!}{%
\begin{tabular}{|l|l|}
\hline \bf Features (\# features) & \bf Accuracy \\ \hline
Glove based feature (200)      & 0.56 \\
Basic (14)                    & 0.59 \\
Grammarly (19)                & 0.96 \\
TF-IDF (100)                  & 0.54 \\
Emotion (4)                   & 0.48 \\
\hline All features (337)     & 0.95 \\
\hline
\end{tabular}}
\label{tab:results_fakenews}
\end{table}

\subsection{Results and Comparisons}
The confusion matrix corresponding to the Celebrity and FakeNewsAMT is given in Figure~\ref{fig:confusion_matrix_celeb} and ~\ref{fig:confusionMatrix_amt}. Tables~\ref{tab:results_celebrity} and ~\ref{tab:results_fakenews} give the 5-fold cross-validation accuracy which we get on Celebrity and FakeNewsAMT datasets, respectively, using individual and combined features. 
Notably, we achieve an accuracy of 78\% and 95\% when using all the features. It entails the fact that our proposed method predicts a high number of news articles correctly. Table~\ref{tab:celebrity_comparison} and ~\ref{tab:fakenewsamt_comparison} compares our results with the recent works \cite{Saikh:20}, \cite{Barua:19}, \cite{Rosas:18}, and \cite{Dabas:18}. Our results are quite competitive.  


\begin{table}
\centering
\caption{Comparison of the proposed method with previous results for the Celebrity dataset}
\resizebox{0.25\textwidth}{!}{%
\begin{tabular}{|c|c|}
\hline
{\bf Model} & {\bf Test Accuracy (\%)}  \\ \hline
Linear SVM \cite{Rosas:18} & 76 \\ \hline
F-NAD \cite{Barua:19} & {\bf 82.61} \\ \hline
Model 1 \cite{Saikh:20}  & 76.53 \\ \hline
Model 2 \cite{Saikh:20}  & 79 \\ \hline
Proposed Work & 78 \\ \hline
\end{tabular}}
\label{tab:celebrity_comparison}
\end{table}

\begin{table}
\centering
\caption{Comparison of the proposed method with previous results for the FakeNewsAMT Dataset}
\resizebox{0.25\textwidth}{!}{%
\begin{tabular}{|c|c|}
\hline
{\bf Model} & {\bf Test Accuracy (\%)}  \\ \hline
Linear SVM \cite{Rosas:18} & 74 \\ \hline
EANN \cite{Dabas:18} & 75.6 \\ \hline
F-NAD \cite{Barua:19} & 81 \\ \hline
Model 1 \cite{Saikh:20} & 77.08 \\ \hline
Model 2 \cite{Saikh:20} & 83.3 \\ \hline
Proposed Work & {\bf 95} \\ \hline
\end{tabular}}
\label{tab:fakenewsamt_comparison}
\end{table}

\subsection{Cross-domain Analysis:}

We perform cross-domain analysis to test how our proposed method helps distinguish fake news across different domains using all the features. We train our best performing classifier on the FakeNewsAMT dataset and test on the Celebrity dataset and vice-versa. Table~\ref{tab:cross_domain_analysis} captures the comparison of results obtained in cross-domain experiment. These results suggest that our method robustly outperforms the previous works.

\begin{table}
\centering
\caption{Results Obtained in Cross-Domain Analysis for best performing classifier}
\label{tab:cross_domain_analysis}
\resizebox{0.45\textwidth}{!}{%
\begin{tabular}{|l|l|l|c|}
\hline
\textbf{Training} & \textbf{Testing} & \textbf{System} & \textbf{Accuracy} \\ \hline
  \multirow{3}{*}{FakeNewsAMT} & \multirow{3}{*}{Celebrity} & Linear SVM \cite{Rosas:18} & 52\% \\ \cline{3-4}
    & & Model 2 \cite{Saikh:20} & 54.3\% \\  \cline{3-4}
    & & Proposed Work &  \textbf{56\%} \\ 
   \hline
  \multirow{3}{*}{Celebrity} &  \multirow{3}{*}{FakeNewsAMT} & Linear SVM \cite{Rosas:18} & 65\% \\ \cline{3-4}
    & & Model 2 \cite{Saikh:20} & 68.5\% \\ \cline{3-4}
    & & Proposed Work &  \textbf{70\%} \\ \hline
\end{tabular}}
\end{table}

\subsection{Learning Curve}
\begin{figure}
  \centering
  \begin{minipage}{0.45\textwidth}
    \includegraphics[width=\textwidth]{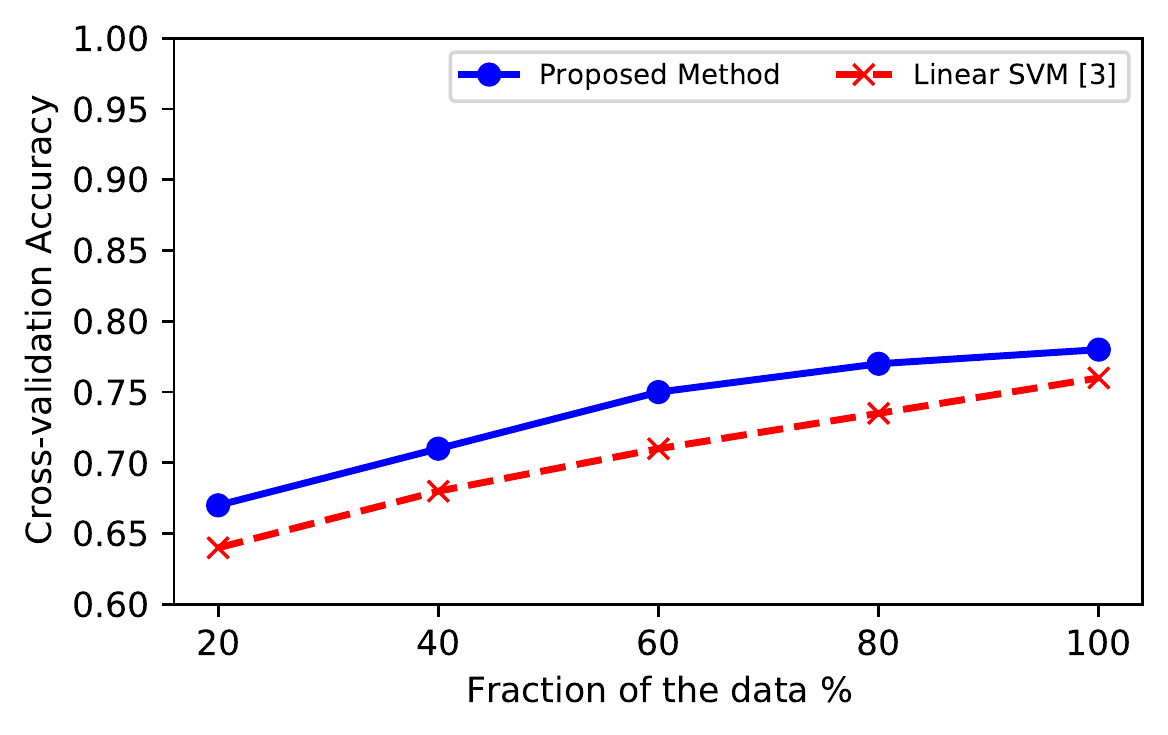}
    \caption{Comparison of the results obtained in Learning Curve experiment with previous work for the Celebrity dataset} \label{fig:learning_curve_celeb}
  \end{minipage}
  \hfill
  \begin{minipage}{0.45\textwidth}
    \includegraphics[width=\textwidth]{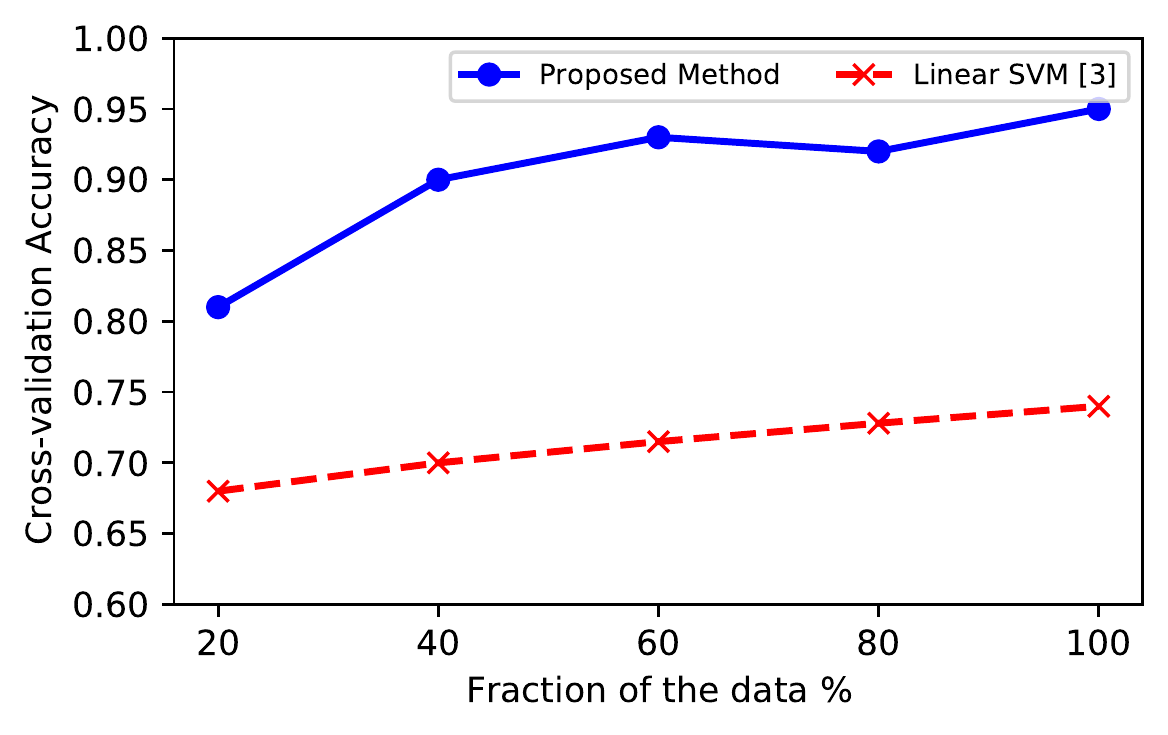}
      \caption{Comparison of the results obtained in Learning Curve experiment with previous work for the FakeNewsAMT dataset} \label{fig:learning_curve_amt}
  \end{minipage}
\end{figure}

We also explore how the amount of data affects the classifier accuracy in identifying fake news. We plot the learning curves for the proposed approach and \cite{Rosas:18} on both the datasets using different fractions of data while training, as shown in figure~\ref{fig:learning_curve_celeb} and ~\ref{fig:learning_curve_amt}. These results suggest (1) our proposed method is outperforming the previous work \cite{Rosas:18}, and (2) our learning curve signifies a steady improvement in both the cases. It implies that a large number of training data has the potential to improve model performance.  

\subsection{Multi-Domain Training and Domain-wise Testing}
FakeNewsAMT dataset contains news articles of six domains (business, education, politics, technology, sports, and entertainment). In this experiment, we train our classifier on the five available domains out of six and test on the remaining domain news articles. Table~\ref{tab:multi_domain} depicts the comparison of results with previous works. It shows that (1) our model outperforms the past results established in this experiment and (2) our model is best performing in the Politics domain and performing worst in the Business domain.  

\begin{table}
\centering
\caption{Comparison of Results Obtained in Multi-domain analysis for best performing classifier}
\label{tab:multi_domain}
\resizebox{0.35\textwidth}{!}{%
\begin{tabular}{|l|l|c|}
\hline
\textbf{Test Domain} & \textbf{Model} & \textbf{Accuracy(\%)} \\ \hline
  \multirow{3}{*}{Technology} & Linear SVM \cite{Rosas:18} & 90 \\ \cline{2-3}
    & Model 2 \cite{Saikh:20} & 88.75 \\ \cline{2-3}
    & Proposed Work &  \textbf{98.7} \\ 
   \hline
  \multirow{3}{*}{Education} & Linear SVM \cite{Rosas:18} & 84 \\ 
  \cline{2-3}
    & Model 2 \cite{Saikh:20} & 91.25 \\ \cline{2-3}
    & Proposed Work &  \textbf{96.2} \\ 
    \hline
    \multirow{3}{*}{Business} & Linear SVM \cite{Rosas:18} & 53 \\ 
  \cline{2-3}
    & Model 2 \cite{Saikh:20} & 78.75 \\ \cline{2-3}
    & Proposed Work &  \textbf{93.7} \\ 
    \hline
    \multirow{3}{*}{Sports} & Linear SVM \cite{Rosas:18} & 51 \\ 
  \cline{2-3}
    & Model 2 \cite{Saikh:20} & 73.75 \\ \cline{2-3}
    & Proposed Work &  \textbf{96.2} \\ 
    \hline
    \multirow{3}{*}{Politics} & Linear SVM \cite{Rosas:18} & 91 \\ 
  \cline{2-3}
    & Model 2 \cite{Saikh:20} & 88.75 \\ \cline{2-3}
    & Proposed Work &  \textbf{100} \\ 
    \hline
    \multirow{3}{*}{Entertainment} & Linear SVM \cite{Rosas:18} & 61 \\ 
  \cline{2-3}
    & Model 2 \cite{Saikh:20} & 76.25 \\ \cline{2-3}
    & Proposed Work &  \textbf{96.2} \\ 
    \hline
\end{tabular}}
\end{table}

\section{Conclusion}
In this paper, we addressed the task of identification of fake news using online tools. We introduced two new sets of features, one obtained through Grammarly and another obtained through Glove-based Feature. Additionally, we introduce the paraphrased article in this problem. Our study shows that the usage of features extracted using Grammarly, Spinbot, and GloVe-based model with other essential features such as Basic, Ngrams, and Emotion Lexicon improves the model performance significantly. The combination of these features achieved the best performance with the Random Forest classifier. Our proposed method obtains a testing accuracy of 78\% on the Celebrity news dataset and 95\% on the FakeNewsAMT dataset.


\bibliographystyle{IEEEtran}

\bibliography{main}

 \newcommand{\noop}[1]{}
\begin{thebibliography}{10}
\providecommand{\url}[1]{#1}
\csname url@samestyle\endcsname
\providecommand{\newblock}{\relax}
\providecommand{\bibinfo}[2]{#2}
\providecommand{\BIBentrySTDinterwordspacing}{\spaceskip=0pt\relax}
\providecommand{\BIBentryALTinterwordstretchfactor}{4}
\providecommand{\BIBentryALTinterwordspacing}{\spaceskip=\fontdimen2\font plus
\BIBentryALTinterwordstretchfactor\fontdimen3\font minus
  \fontdimen4\font\relax}
\providecommand{\BIBforeignlanguage}[2]{{%
\expandafter\ifx\csname l@#1\endcsname\relax
\typeout{** WARNING: IEEEtran.bst: No hyphenation pattern has been}%
\typeout{** loaded for the language `#1'. Using the pattern for}%
\typeout{** the default language instead.}%
\else
\language=\csname l@#1\endcsname
\fi
#2}}
\providecommand{\BIBdecl}{\relax}
\BIBdecl

\bibitem{Mihalcea:09}
R.~Mihalcea and C.~Strapparava, ``The lie detector: Explorations in the
  automatic recognition of deceptive language,'' in \emph{Proceedings of the
  ACLIJCNLP {2009} Conference Short Papers}, 2009.

\bibitem{Ott:11}
M.~Ott, Y.~Choi, C.~Cardie, and J.~T. Hancock, ``Finding deceptive opinion spam
  by any stretch of the imagination,'' in \emph{Proceedings of ACL}, 2011, pp.
  309--319, version 1.

\bibitem{Rosas:18}
V.~Pérez-Rosas, B.~Kleinberg, A.~Lefevre, and R.~Mihalcea, ``Automatic
  detection of fake news,'' in \emph{Proceedings of the 27th International
  Conference on Computational Linguistics. Association for Computational
  Linguistics}.\hskip 1em plus 0.5em minus 0.4em\relax Santa Fe, New Mexico,
  USA: Association for Computational Linguistics, 2017, p. 3391–3401.

\bibitem{Rubin:16}
V.~L. Rubin, N.~J. Conroy, Y.~Chen, and S.~Cornwell, ``Fake news or truth?
  using satirical cues to detect potentially misleading news,'' in
  \emph{Proceedings of NAACL-HLT}, San Diego, California, 2016, pp. 7--17.

\bibitem{NYUTSE:20}
{NYU Tandon School of Engineering}, ``Red-flagging misinformation could slow
  the spread of fake news on social media,'' Avaliable at
  \url{https://www.sciencedaily.com/releases/2020/04/200428112542.htm}
  (2020/28/4).

\bibitem{Granik:17}
M.~Granik and V.~Mesyura, ``Fake news detection using naive bayes classifier,''
  in \emph{Electrical and Computer Engineering (UKRCON), 2017 IEEE First
  Ukraine Conference on}.\hskip 1em plus 0.5em minus 0.4em\relax IEEE, 2017,
  pp. 900--903.

\bibitem{goyalgait}
D.~{Goyal}, K.~R. {Jerripothula}, and A.~{Mittal}, ``Detection of gait
  abnormalities caused by neurological disorders,'' in \emph{2020 IEEE 22nd
  International Workshop on Multimedia Signal Processing (MMSP)}, 2020, pp.
  1--6.

\bibitem{Castillo:11}
C.~Castillo, M.~Mendoza, and B.~Poblete, ``Information credibility on
  twitter,'' in \emph{Proceedings of the 20th International Conference on World
  Wide Web}.\hskip 1em plus 0.5em minus 0.4em\relax ACM, 2011, pp. 675--684.

\bibitem{Feng:12}
S.~Feng, R.~Banerjee, and Y.~Choi, ``Syntactic stylometry for deception
  detection,'' in \emph{Proceedings of the 50th Annual Meeting of the
  Association for Computational Linguistics: Short Papers-Volume 2}.\hskip 1em
  plus 0.5em minus 0.4em\relax Association for Computational Linguistics, 2012,
  pp. 171--175.

\bibitem{Qazvinian:11}
V.~Qazvinian, E.~Rosengren, D.~R. Radev, and Q.~Mei, ``{Rumor has it:}
  identifying misinformation in microblogs,'' in \emph{Proceedings of the
  Conference on Empirical Methods in Natural Language Processing}.\hskip 1em
  plus 0.5em minus 0.4em\relax Association for Computational Linguistics, 2011,
  pp. 1589--1599.

\bibitem{Takahashi:12}
T.~Takahashi and N.~Igata, ``Rumor detection on twitter,'' in \emph{Proceedings
  of the 6th International Conference on Soft Computing and Intelligent
  Systems, and the 13th International Symposium on Advanced Intelligence
  Systems}, 2012, pp. 452--457.

\bibitem{Gupta:14}
A.~Gupta, P.~Kumaraguru, C.~Castillo, and P.~Meier, ``Tweetcred: Real-time
  credibility assessment of content on twitter,'' in \emph{International
  Conference on Social Informatics}.\hskip 1em plus 0.5em minus 0.4em\relax
  Springer, 2014, pp. 228--243.

\bibitem{Popat:16}
K.~Popat, S.~Mukherjee, J.~Strötgen, and G.~Weikum, ``Credibility assessment
  of textual claims on the web,'' in \emph{Proceedings of the 25th ACM
  International on Conference on Information and Knowledge Management}.\hskip
  1em plus 0.5em minus 0.4em\relax ACM, 2016, pp. 2173–--2178.

\bibitem{Shu:18}
K.~Shu, H.~R. Bernard, and H.~Liu, ``Studying fake news via network analysis:
  Detection and mitigation,'' \emph{CoRR}, 2018.

\bibitem{Zhao:15}
Z.~Zhao, P.~Resnick, and Q.~Mei, ``{Enquiring Minds:} early detection of rumors
  in social media from enquiry posts,'' in \emph{Proceedings of the 24th
  International Conference on World Wide Web}.\hskip 1em plus 0.5em minus
  0.4em\relax International World Wide Web Conferences Steering Committee,
  2015, pp. 1395--1405.

\bibitem{Buntain:2017}
C.~Buntain and J.~Golbeck, ``Automatically identifying fake news in popular
  twitter thread,'' in \emph{Smart Cloud (SmartCloud), 2017 IEEE International
  Conference}.\hskip 1em plus 0.5em minus 0.4em\relax IEEE, 2017, pp. 208--215.

\bibitem{Shu:19}
K.~Shu, A.~Sliva, S.~Wang, J.~Tang, and H.~Liu, ``Fake news detection on social
  media: A data mining perspective,'' \emph{ACM SIGKDD Explorations Newsletter
  19}, pp. 22--36, 2019.

\bibitem{Khattar:19}
D.~Khattar, G.~J. Singh, M.~Gupta, and V.~Varma, ``{MVAE:} multimodal
  variational autoencoder for fake news detection,'' in \emph{Proceedings of
  the 2019 World Wide Web Conference}.\hskip 1em plus 0.5em minus 0.4em\relax
  ACM, 2019.

\bibitem{10.1007/978-3-319-46478-7_12}
K.~R. Jerripothula, J.~Cai, and J.~Yuan, ``Cats: Co-saliency activated tracklet
  selection for video co-localization,'' in \emph{Computer Vision -- ECCV
  2016}, B.~Leibe, J.~Matas, N.~Sebe, and M.~Welling, Eds.\hskip 1em plus 0.5em
  minus 0.4em\relax Cham: Springer International Publishing, 2016, pp.
  187--202.

\bibitem{shah2017multimodal}
R.~Shah and R.~Zimmermann, \emph{Multimodal analysis of user-generated
  multimedia content}.\hskip 1em plus 0.5em minus 0.4em\relax Springer, 2017.

\bibitem{8290832}
K.~R. {Jerripothula}, J.~{Cai}, and J.~{Yuan}, ``Efficient video object
  co-localization with co-saliency activated tracklets,'' \emph{IEEE
  Transactions on Circuits and Systems for Video Technology}, vol.~29, no.~3,
  pp. 744--755, 2019.

\bibitem{Rashkin:17}
H.~Rashkin, E.~Choi, J.~Y. Jang, S.~Volkova, and Y.~Choi, ``Truth of varying
  shades: Analyzing language in fake news and political fact-checking,'' in
  \emph{Proceedings of the 2017 Conference on Empirical Methods in Natural
  Language Processing}.\hskip 1em plus 0.5em minus 0.4em\relax Association for
  Computational Linguistics, 2017, pp. 2931--2937.

\bibitem{Martino:19}
G.~D.~S. Martino, S.~Yu, A.~Barrón-Cedeño, R.~Petrov, and P.~Nakov,
  ``Fine-grained analysis of propaganda in news articles,'' in
  \emph{Proceedings of the 2019 Conference on Empirical Methods in Natural
  Language Processing and the 9th International Joint Conference on Natural
  Language Processing (EMNLP-IJCNLP)}.\hskip 1em plus 0.5em minus 0.4em\relax
  Association for Computational Linguistics, 2019, p. 5636–5646.

\bibitem{shivangi:19}
S.~Singhal, R.~R. Shah, T.~Chakraborty, P.~Kumaraguru, and S.~Satoh,
  ``Spotfake: A multi-modal framework for fake news detection,'' 2019.

\bibitem{parikh:19}
S.~B. Parikh and P.~K. Khedia, Saurin R.and~Atrey, ``A framework to detect fake
  tweet images on social media,'' 2019.

\bibitem{8269367}
K.~R. {Jerripothula}, J.~{Cai}, and J.~{Yuan}, ``Quality-guided fusion-based
  co-saliency estimation for image co-segmentation and colocalization,''
  \emph{IEEE Transactions on Multimedia}, vol.~20, no.~9, pp. 2466--2477, 2018.

\bibitem{8099896}
K.~R. {Jerripothula}, J.~{Cai}, J.~{Lu}, and J.~{Yuan}, ``Object
  co-skeletonization with co-segmentation,'' in \emph{2017 IEEE Conference on
  Computer Vision and Pattern Recognition (CVPR)}, 2017, pp. 3881--3889.

\bibitem{singhal2019spotfake}
S.~Singhal, R.~R. Shah, T.~Chakraborty, P.~Kumaraguru, and S.~Satoh,
  ``Spotfake: A multi-modal framework for fake news detection,'' in \emph{2019
  IEEE Fifth International Conference on Multimedia Big Data (BigMM)}.\hskip
  1em plus 0.5em minus 0.4em\relax IEEE, 2019, pp. 39--47.

\bibitem{singhal2020spotfake+}
S.~Singhal, A.~Kabra, M.~Sharma, R.~R. Shah, T.~Chakraborty, and P.~Kumaraguru,
  ``Spotfake+: A multimodal framework for fake news detection via transfer
  learning (student abstract).'' in \emph{AAAI}, 2020, pp. 13\,915--13\,916.

\bibitem{7484309}
K.~R. {Jerripothula}, J.~{Cai}, and J.~{Yuan}, ``Image co-segmentation via
  saliency co-fusion,'' \emph{IEEE Transactions on Multimedia}, vol.~18, no.~9,
  pp. 1896--1909, 2016.

\bibitem{grammarly:19}
S.~Price, ``How can grammarly help you to write correct blog post?'' Available
  at
  \url{https://blog.hubspot.com/website/how-can-grammarly-help-write-correct-blog-post}
  (2019/9/27).

\bibitem{macqueen1967some}
J.~MacQueen \emph{et~al.}, ``Some methods for classification and analysis of
  multivariate observations,'' in \emph{Proceedings of the fifth Berkeley
  symposium on mathematical statistics and probability}, vol.~1, no.~14.\hskip
  1em plus 0.5em minus 0.4em\relax Oakland, CA, USA, 1967, pp. 281--297.

\bibitem{7025663}
K.~R. {Jerripothula}, J.~{Cai}, F.~{Meng}, and J.~{Yuan}, ``Automatic image
  co-segmentation using geometric mean saliency,'' in \emph{2014 IEEE
  International Conference on Image Processing (ICIP)}, 2014, pp. 3277--3281.

\bibitem{7457899}
K.~R. {Jerripothula}, J.~{Cai}, and J.~{Yuan}, ``Qcce: Quality constrained
  co-saliency estimation for common object detection,'' in \emph{2015 Visual
  Communications and Image Processing (VCIP)}, 2015, pp. 1--4.

\bibitem{jerripothula2017co}
K.~R. Jerripothula, ``Co-saliency based visual object co-segmentation and
  co-localization,'' Ph.D. dissertation, Nanyang Technological University,
  2017.

\bibitem{jerripothula2015group}
K.~R. Jerripothula, J.~Cai, and J.~Yuan, ``Group saliency propagation for large
  scale and quick image co-segmentation,'' in \emph{2015 IEEE International
  Conference on Image Processing (ICIP)}.\hskip 1em plus 0.5em minus
  0.4em\relax IEEE, 2015, pp. 4639--4643.

\bibitem{Jones:04}
K.~S. Jones, ``Idf term weighting and ir research lessons,'' \emph{Journal of
  Documentation}, pp. 521--523, 2004.

\bibitem{textblob:18}
S.~Jain, ``Natural language processing for beginners: Using textblob,''
  Available at
  \url{https://www.analyticsvidhya.com/blog/2018/02/natural-language-processing-for-beginners-using-textblob/}
  (2018/2/11).

\bibitem{9153789}
K.~R. {Jerripothula}, A.~{Rai}, K.~{Garg}, and Y.~S. {Rautela}, ``Feature-level
  rating system using customer reviews and review votes,'' \emph{IEEE
  Transactions on Computational Social Systems}, pp. 1--10, 2020.

\bibitem{Saikh:20}
T.~Saikh, A.~De, A.~Ekbal, and P.~Bhattacharyya, ``A deep learning approach for
  automatic detection of fake news,'' 2020.

\bibitem{Barua:19}
R.~Barua, R.~Maity, D.~Minj, T.~Barua, and A.~K. Layek, ``{F-NAD:} an
  application for fake news article detection using machine learning
  techniques,'' in \emph{IEEE Bombay Section Signature Conference (IBSSC)},
  India, 2019.

\bibitem{Dabas:18}
K.~Dabas, P.~Kumaraguru, T.~Chakraborty, and R.~R. Shah, ``Multimodal fake news
  detection on online socials,'' Master's thesis, IIITD, 2018.

\end{thebibliography}

\end{document}